\def\year{2022}\relax
\documentclass[letterpaper]{article} 
\usepackage{aaai22}  
\usepackage{times}  
\usepackage{helvet}  
\usepackage{courier}  
\usepackage[hyphens]{url}  
\usepackage{graphicx} 
\usepackage{natbib}  
\usepackage{caption} 
\DeclareCaptionStyle{ruled}{labelfont=normalfont,labelsep=colon,strut=off} 
\usepackage{algorithm}
\usepackage{algorithmic}
\usepackage{xcolor}
\newcommand{\answerYes}[1]{\textcolor{blue}{#1}} 
\newcommand{\answerNo}[1]{\textcolor{teal}{#1}} 
\newcommand{\answerNA}[1]{\textcolor{gray}{#1}}

\usepackage{newfloat}
\usepackage{listings}
\floatstyle{ruled}
\newfloat{listing}{tb}{lst}{}
\floatname{listing}{Listing}
\title{Combining Objective and Subjective Perspectives for Political News Understanding}
\author{
Evan Dufraisse,\textsuperscript{\rm 1}\textsuperscript{\rm 2}
 Adrian Popescu,\textsuperscript{\rm 1}
 Julien Tourille,\textsuperscript{\rm 1}
 Armelle Brun,\textsuperscript{\rm 2}
 Olivier Hamon,\textsuperscript{\rm 3}
}

\affiliations{%
    \textsuperscript{\rm 1}Université Paris-Saclay, CEA, List, F-91120, Palaiseau, France\\
    \textsuperscript{\rm 2}Université de Lorraine - CNRS - Loria, Vandoeuvre les Nancy Cedex, France\\
    \textsuperscript{\rm 3}Syllabs, Paris, France\\
    evan.dufraisse@cea.fr, adrian.popescu@cea.fr, julien.tourille@cea.fr, armelle.brun@loria.fr, hamon@syllabs.com
}
\usepackage{enumitem}
\usepackage{xcolor}
\usepackage{multirow}
\usepackage{tabularx}

\begin{document}

\maketitle

\begin{abstract}
Researchers and practitioners interested in computational politics rely on automatic content analysis tools to make sense of the large amount of political texts available on the Web.
Such tools should provide objective and subjective aspects at different granularity levels to make the analyses useful in practice. 
Existing methods produce interesting insights for objective aspects, but are limited for subjective ones, are often limited to national contexts, and have limited explainability. 
We introduce a text analysis framework which integrates both perspectives and provides a fine-grained processing of subjective aspects.
Information retrieval techniques and knowledge bases complement powerful natural language processing components to allow a flexible aggregation of results at different granularity levels.   
Importantly, the proposed bottom-up approach facilitates the explainability of the obtained results.
We illustrate its functioning with insights on news outlets, political orientations, topics, individual entities, and demographic segments. 
The approach is instantiated on a large corpus of French news, but is designed to work seamlessly for other languages and countries.
\end{abstract}

\section{Introduction}
Political news provides essential information, the interpretation of which shapes our understanding of political actions and events.
Analyzing the vast amount of political news available on the web is only possible by automating the process.
In order to maximize its impact for stakeholders in computational politics, such analysis should be comprehensive and flexible.  
Most news analysis tools focus on objective metrics, or include only very coarse subjective metrics. 
Textmap~\cite{lloyd2005lydia} detects entities and their relationships, and grounds them in space and time. \citet{godboleLargeScaleSentimentAnalysis, plochAdvancedPressReview2016, bautinInternationalSentimentAnalysis2021} extend Textmap's purely objective analysis using lexicon-based sentiment analysis at the article level, which does not allow for precise sentiment analysis towards article entities.
A citation graph analysis allows the clustering of news outlets based on their political affinity~\cite{cointet2021uncovering}.
An aggregation of TV and radio broadcast metadata is used to estimate political biases~\cite{hosting_cag_2022}.

More recently, the quantification of subjective aspects has also been approached with a focus on classifying the political leanings of entire news articles using large language models~\cite{baly2020we,ko2023khan,kamal2023learning}.
These approaches have important shortcomings, notably in terms of granularity, explainability, and the need to compile a new dataset when the political context changes.  

We tackle these limitations by introducing a news analysis framework which provides comprehensive objective and subjective insights.
These insights can be aggregated in a flexible manner to provide insights about news outlets, political topics, individual entities and demographic segments. 
To obtain this result, the framework integrates recent NLP components, information retrieval techniques, news outlet metadata, temporal metadata and external knowledge bases.
In particular, we deploy target-dependent sentiment classification (TSC) to obtain a fine-grained subjective representation of articles.
The framework is designed to work seamlessly for multiple countries since all NLP components are multilingual and the knowledge bases are not country specific. 
It is instantiated for political news, using a French political news corpus.
The main findings are:
\begin{itemize}[leftmargin=*,noitemsep,topsep=0pt]
\item mainstream political orientations are presented in a rather balanced way in the major news outlets, but the radical left and right are positively and negatively biased, respectively;
\item the mentions and sentiments associated with political orientations vary across impactful political topics;
\item sentiment scores are generally negative, with important variation between news outlets;
\item the most positive and negative sentiment scores for politicians are correlated with the public perception of their actions;
\item mentions are biased toward male politicians, but sentiment scores of female politicians are higher;
\item there is an age bias toward older politicians;
\item the French semi-presidential system is reflected in the news, with a dominance of presidents or presidential candidates.
\end{itemize}

These findings can be used by multiple stakeholders. 
News outlets can analyze their positioning and make it more transparent to the general public.
Social scientists can gain new insights into the online representation of the political landscape.
Political organizations can monitor the online reporting of their political actions.
Citizens can be informed about potential political, topical and demographic biases of news outlets.

\section{Related Work}
Media analysis studies are based on tools and insights from different disciplines, including political science~\cite{measuring_kim_2022} economics~\cite{10.1257/aer.20160812}, and NLP~\cite{agrawal-etal-2022-towards,gangula-etal-2019-detecting,spinde-etal-2021-neural-media}.
Research efforts that reveal political, thematic or demographic biases are of particular interest.

Gender representation has been shown to be biased in multiple countries. 
\citet{electoral_thesen_2022} show that male politicians are more present in news reporting for countries with proportional representation. 
\citet{dacon2021} analyze news abstracts and conclude to an underrepresentation of women in English news.
\citet{wevers-2019-using} and \citet{bastin_gender_2022} investigate gender bias in Dutch and French news, respectively. 
Online gender bias tracking tools were also developed, such as Gender Gap Tracker~\cite{asr2021gender} for Canada and Gendered News~\cite{richard2022genderednews} for France.

Aside from gender, topic-related bias in local newspapers is investigated in~\citet{rivas_eu}, with focus on the European Union.
\citet{spinde2020_integrated} study the bias of words in German news.
\citet{hosting_cag_2022} show that political orientations are unevenly represented across broadcast outlets by aggregating mentions of individual political and non-political figures.
\citet{cointet2021uncovering} analyze the French media ecosystem by considering the online interaction graph between them. 
These studies are interesting for quantifying mentions but do not address the measurement of subjectivity. 

Framing~\cite{entmanFramingClarificationFractured1993}, the action of emphasizing certain facts over others, has also been studied. \citet{liu-etal-2019-detecting} introduce the  Gun Violence Frame Corpus and implement an approach for news framing detection. Their results show that framing is dynamic and follows closely the gun-related violent crimes. \citet{coverage_young_2021} investigate how immigration is framed in US newspapers over the last decade, and show an increase in negative messages about immigrants. \citet{hamborg-2020-media} studies framing through sentiment classification and word choice. The author implements a method for highlighting media outlet position over a specific topic.
\citet{alonso2023framing} very recently tested large language models for frames detection in news headlines. Stance detection~\cite{kuccuk2020stance}, i.e. the subjective positioning of the author towards a target, is also actively investigated. \citet{conforti-etal-2020-stander} build a corpus of news focused on company merging stance detection. \citet{torres_stance} address the detection of misleading news headlines through stance comparison of the body and the text. Other approaches treat stance detection as a text-level sentiment analysis task~\cite{godboleLargeScaleSentimentAnalysis, plochAdvancedPressReview2016,wu-etal-2020-sentirec,bautinInternationalSentimentAnalysis2021}, but as \citet{bestvater_monroe_2023} points out, this level is too coarse to bring satisfactory insights about stance. 

More recently, some approaches \cite{baly2020we,ko2023khan,kamal2023learning} have used stance-labeled article databases such as AllSides to train language models for stance-classification in an end-to-end fashion. As briefly stated in introduction those approaches have several shortcomings: (1) stance is characterized at a coarse level of granularity, and it is impossible to provide detailed insights about politicians, political organizations, or demographic segments;
(2) the explanatory power of the results obtained is limited, since only one class label per article is provided;
and
(3) the task is country and time-dependent and relies on supervised annotated datasets, thus requiring the creation of new datasets for each different political context.

Although they provide useful analytical tools for capturing media bias, most studies use lexical, mention-count based, or metadata approaches that fail to account for the sentiment expressed toward key entities in the text.
This is mainly an effect of the relative lack of TSC resources for the political domain. 
Target-dependent Sentiment Classification~(TSC) predicts the opinion towards a precise entity and can be aggregated to better understand the positioning of a news outlet toward that entity~\cite{Hamborg2023}. 
TSC is often evaluated on short texts such as tweets \cite{nakov-etal-2016-semeval}, reviews~\cite{pontiki-etal-2016-semeval} or comments~\cite{severyn2016multi}. 
However, news texts are more challenging because sentiments are expressed implicitly or indirectly~\cite{hamborg-donnay-2021-newsmtsc}, often include multiple targets in a single sentence~\cite{brauwers2022survey}, and both negative and positive arguments about a target entity are combined due to the fact that journalists are supposed to be objective~\cite{balahur-etal-2010-sentiment,hamborg2019automated,liu2010sentiment}. 
\citet{hamborg-donnay-2021-newsmtsc} introduced a new dataset of news sentences extracted from US newspapers annotated with polarity at the entity level. 
\citet{dufraisseMADTSCMultilingualAligned2023} address multilingual target sentiment analysis in news by creating an aligned dataset in several European languages.
TSC resources and methods such as the ones described~\cite{brauwers2022survey,dufraisseMADTSCMultilingualAligned2023} are central to our work, and we integrate them into our framework.

\section{News Processing Framework}
\label{section:pipeline}
This section presents the main technical components of the proposed news analysis framework.

\noindent\textbf{Corpus collection.}
The corpus was built by collecting RSS streams from more than~280 politically diverse French news outlets between January~2016 to December~2022. 
Texts of the articles were extracted from their HTML using Trafilatura~\cite{barbaresi-2021-trafilatura}. 
As RSS were not completely mutually exclusive, and because some articles may be available from multiple links due to revision versions, a near deduplication using MinHashLsh~\cite{broder1998} was performed to only keep one version per domain name. 
Parameters are provided in Appendix~\ref{app:parameters}.
Only news including 200 characters or more were retained, totaling over 457,000 articles.
Corpus statistics are provided in Appendix~\ref{app:statistics}.

\noindent\textbf{Entity detection.}
Articles were split into sentences using the French ``fr\_web\_core\_sm'' Spacy model.
We then used the French version of Flair~\cite{akbik-etal-2019-flair} for entity detection.
There are 2.27M mentions of persons in the corpus.

\noindent\textbf{Entity linking.}
To solve ambiguities, entities were linked to Wikidata~\cite{pellissier2016freebase} using mGenre~\cite{de-cao-etal-2022-multilingual}, a recent multilingual linking model. 
Linking was considered successful if the log-likelihood of the best candidate was greater than -0.2.
The threshold was determined by labeling 200 predictions per 0.1 range between 0 and 0.5, and it provides over 95\% accuracy. There are 1.27M linked persons left.

\noindent\textbf{Use of knowledge bases.}
Wikidata~\cite{pellissier2016freebase} entries of linked persons were used to select politicians and extract their gender, birth date, country, and political affiliation(s).
ParlGov~\cite{doring2012parliament} includes information about political parties in different countries. 
It uses a scale from 0 to 10 (left to right) to encode political orientation.
There are 520K mentions of politicians (89.9K female), out of which 345K are French and 333K can be linked to political orientations from ParlGov.
To facilitate interpretability, we reduce this to a 5 points scale by merging pairs of neighboring orientations. 
The resulting scale used in experiments includes the following orientations: RL - radical left (30.5K mentions); CL - center left (56.4K); C - center (113.3K); CR - center right (87.3K); RR - radical right (45.8K).
We finally match French political party names from Wikidata and ParlGov to attribute an orientation to politicians. 

\noindent\textbf{Target-dependent sentiment classification}
 aims to determine the sentiment that is expressed towards a given entity in a given context. 
TSC resources for the political domain were only recently made available for English~\cite{hamborg-donnay-2021-newsmtsc} and for multiple languages~\cite{dufraisseMADTSCMultilingualAligned2023}.
We use the French version of this last dataset to train an SPC-Bert model~\cite{song2019attentional}, which is based on the classical Sentence Pair Classification task of Bert~\cite{devlin-etal-2019-bert} and provides strong TSC performances~\cite{hamborg-donnay-2021-newsmtsc,dufraisseMADTSCMultilingualAligned2023}.
The underlying transformer architecture is CamemBert~\cite{martin-etal-2020-camembert}, the French equivalent of RoBERTa~\cite{liu2019roberta}. 
We follow the training procedure described in~\cite{dufraisseMADTSCMultilingualAligned2023} and produce inferences for negative, neutral, and positive classes. The training procedure uses a standard train/dev/test split, with each split representing 70\%/10\%/20\% of the entries respectively. 
The model achieves an $F1_{macro}$ score of $70.8$.
It is important to assess the impact of errors made by the TSC component.
We run a preliminary experiment with two samples. A first, for which we keep all sentiment predictions, and a second, using a threshold, for which we only keep the 50\% most confident ones for each class. 
For both samples, we compute the average sentiment scores and the number of mentions for the 1,000 most cited politicians from the corpus.
We then compute the Pearson correlation between the sampled distributions, and obtain values of 0.993 and 0.966 for the number of mentions and the sentiments distributions, respectively.
These results show that, while individual predictions are imperfect, their aggregation leads to very stable results, and all sentiment predictions are usable.

\begin{figure*}[ht]
	\centering
	\includegraphics[width=0.99\linewidth, trim={0cm 0cm 0cm 0cm}]{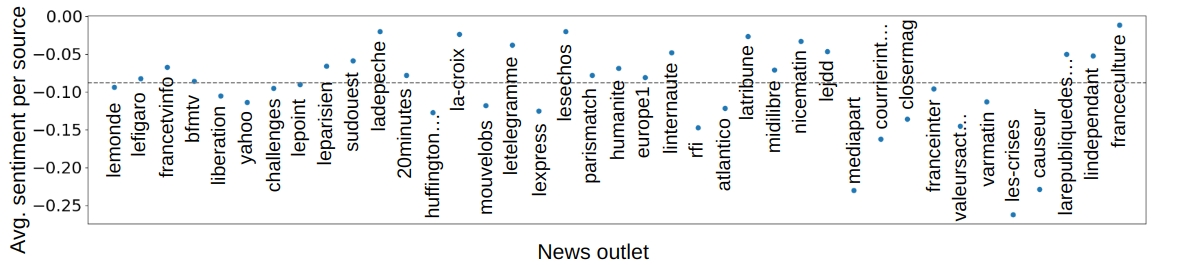} 
\vspace{-3mm}
	\caption{Average sentiment (blue dots) for the most frequent 40 news outlets in the corpus. The dashed line represents the average sentiment for the entire corpus. Outlet frequency decreases from left to right.
	}
	\label{fig:source_sentiment}
\end{figure*}

\noindent\textbf{Topic analysis.}
We use an information retrieval approach to select relevant subsets of documents for political topics. 
This type of approach is interesting because it enables an easy and flexible definition of topics.
The news corpus was indexed using BM25~\cite{robertson2009probabilistic}, a probabilistic text representation method which provides strong performance when compared to more recent neural-based approaches~\cite{deshmukh2020}, \cite{khloponin2021}.
We chose a set of 10 impactful political topics associated with queries for retrieval. 
The set includes both diversified topics and similar ones. 
The list of topics includes:
(1) climate change, 
(2) corruption in the political domain,
(3) the economic consequences of the Covid-19 crisis,
(4) the health-related issues of Covid-19,
(5) the yellow vests protest which took place in France,
(6) immigration,
(7) purchasing power,
(8) the war in Syria,
(9) the war in Ukraine
and (10) the economic consequences of the war in Ukraine.
These topics are used in the analysis of the Subsection~\ref{subsec:topic}.
Pertinence thresholds are needed to select relevant subsets of documents for each topic.
These were determined empirically by examining the titles and snippets of BM25 returned results.
There are between 2.1K and 8K mentions per topic in the corpus.

\section{Political News Analysis}
\label{sec:analysis}
We propose a comprehensive analysis of political representation in the news by combining objective and subjective perspectives.
We analyze news outlets, political topics, politicians, and demographic segments (gender, age).

\subsection{Outlet-oriented Analysis}
\label{subsec:source}
\noindent\textbf{Source Sentiment}. 
Figure~\ref{fig:source_sentiment} presents the average sentiment for mentions of politicians for the most frequent news outlets in the corpus. 
A short description of each source mentioned below is proposed in Appendix~\ref{app:source_description}.
The political leanings of sources are those established in~\citet{cointet2021uncovering}. 
The average sentiment across all outlets is negative with an average of -0.083, but with significant variability between sources, which can be explained by various factors such as their nature, geographic focus and/or political leanings.
The lowest scores are obtained for \textit{les-crises, causeur} and \textit{mediapart}, which is consistent with their characterization by experts as right-wing anti-establishment; right-wing and left-wing outlets, respectively.
High scores are obtained for regional or local news outlets such as \textit{la depeche}, \textit{nicematin} or \textit{letelegramme}, as well as for economic media such as \textit{lesechos} or \textit{latribune}.
As for the major outlets, their scores are close to the average, with \textit{lefigaro} being the least negative, followed by \textit{lemonde} and \textit{liberation}.

\begin{figure*}[ht]
	\centering
	\includegraphics[width=0.999\linewidth, trim={0cm 0cm 0cm 0cm}]{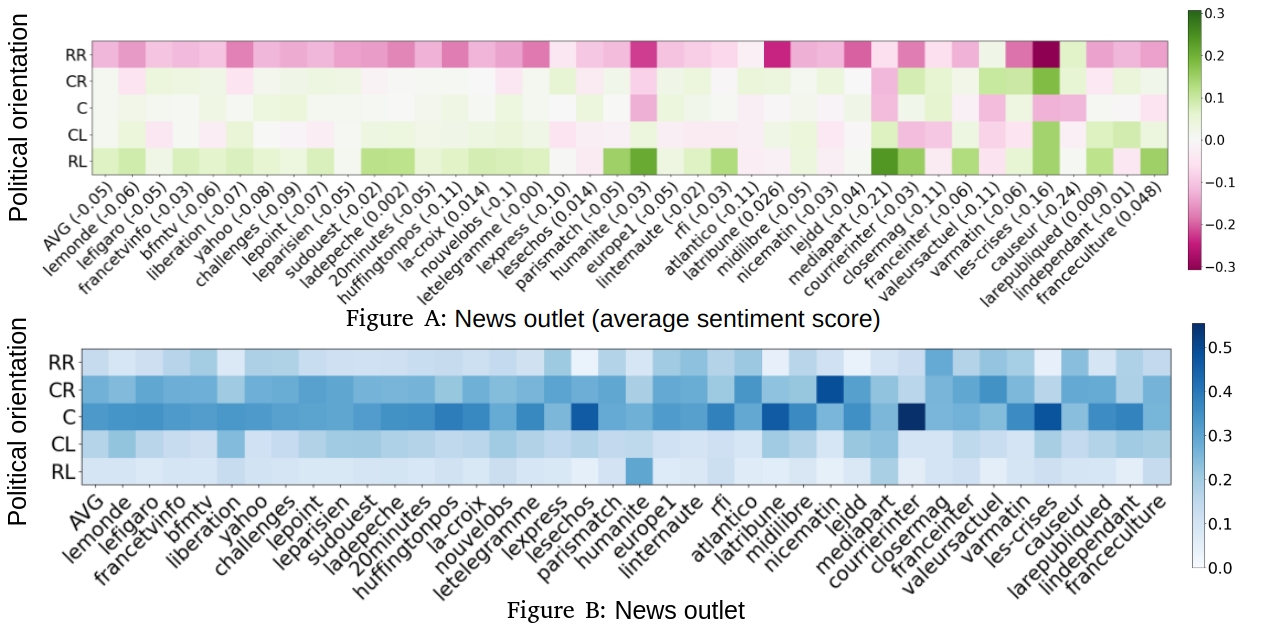} 
 \vspace{-5mm}
	\caption{\textbf{Fig. A}: Deviation of the sentiment associated with major political orientations from the average sentiment of each source. Average sentiment of linked politicians is indicated in parentheses for each source. \textbf{Fig. B}: Distribution of mentions of political orientations in news outlets.
	}
	\label{fig:mention_positioning}
\end{figure*}


\noindent\textbf{Mapping of Political Orientations}. 
We analyze the positioning of news outlets with respect to major political orientations by aggregating mentions and sentiment scores.
The mention-oriented analysis is similar to existing ones~\cite{hosting_cag_2022}, while the sentiment-oriented analysis adds a subjective perspective.
We use the five-points political orientation mapping described in Section~\ref{section:pipeline}.

Figure~\ref{fig:mention_positioning} summarizes the mappings of mentions and of sentiment scores \textit{vs.} political orientations.
There is no correlation between mentions and sentiment since the Pearson correlation coefficient between the AVG columns in the two figures is 0.06. 
This result shows that the two types of analyses are complementary and provide additional insights into political news. 
Right-leaning orientations are more cited than their left-leaning counterparts but the sentiment associated with right-leaning orientations is more negative. 
This contrast is even clearer for RR and RL, the two radical orientations that are represented in the two figures.
A complementary analysis per news outlet is presented in Appendix~\ref{app:source_ms}.

Figure~\ref{fig:mention_positioning} shows that the center, including the French governing party, is the most mentioned orientation.
The right-wing tendencies are more represented than their left-wing counterparts. 
Outlet-wise, the center is overrepresented in economic newspapers (\textit{latribune} and \textit{lesechos}), but also in \textit{les-crises}, a right-wing anti-establishment outlet. 
RL is strongly present in \textit{humanite}, the newspaper of the French Communist Party, while the extreme right is often cited by sources such as ~\textit{closermag}, a tabloid, or \textit{causeur}, a right-wing outlet. 

Figure~\ref{fig:mention_positioning} illustrates the sentiment-oriented positioning of news outlets.
It is quantified by the difference between the sentiment score per orientation and the average sentiment score of the source. 
The positioning of sources toward political orientations varies, and this is a positive finding since diversified opinions are required for a healthy democratic debate. 
The representation of mainstream orientations (CL, C, CR) is rather balanced in the major sources (left of the figure) and more variable in other source (right of the figure).
We note that, overall, there is a positive bias toward radical left (RL) politicians and a negative bias toward radical right (RR) ones. 
Notable exceptions for RL include \textit{valeursactuelles}, \textit{closermag} and \textit{lindependant}.
The only two sources which have a slightly positive positioning toward RR relative to their average positioning are \textit{valeursactuelles} and \textit{causeur}, but the average sentiment scores of RR remain overall negative even for these two sources.
Intuitively, the center, which includes the current ruling party, is most criticized by news outlets which are left-wing (\textit{humanite} and \textit{mediapart}) and right-wing (\textit{valeursactuelles} and \textit{causeur}).
A complementary analysis of the temporal variation of mentions of political orientations and of sentiment associated with them is presented in Appendix~\ref{app:temporal_orientations}. 

We also note that the average sentiment of sources differs between Figures~\ref{fig:source_sentiment} and~\ref{fig:mention_positioning}.
The scores obtained for French-only politicians are higher than when including all politicians.
This may be explained by a more critical stance of French media on international issues than on national ones.

\subsection{Topic Oriented Analysis} 
\label{subsec:topic}
\begin{figure}
	\centering
	\includegraphics[width=0.999\linewidth, trim={0cm 0cm 0cm 0cm}]{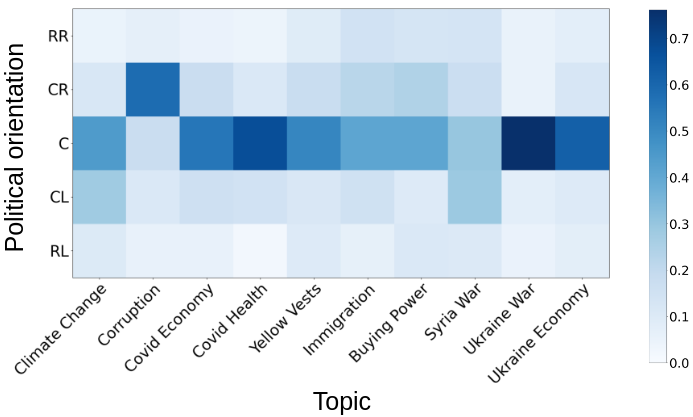} 
 \vspace{-4mm}
	\caption{Distribution of mentions of political orientations for ten impactful political topics.}
	\label{fig:topics_mentions}
\end{figure}

\begin{figure}
	\centering
	\includegraphics[width=0.999\linewidth, trim={0cm 0cm 0cm 0cm}]{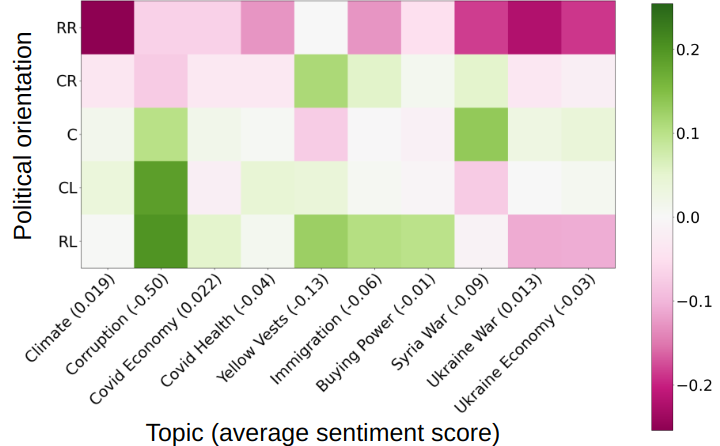} 
 \vspace{-4mm}
	\caption{Deviation of the sentiment for major political orientations from the average sentiment of each topic. Average topic sentiment is given in parentheses. 
	}
	\label{fig:topics_sentiment}
\end{figure}

The results for mentions and sentiments associated with political topics are presented in Figures~\ref{fig:topics_mentions} and~\ref{fig:topics_sentiment}.
The mentions of centrist orientation, which includes the governing party, dominate most topics.
This is particularly the case for the health effects of Covid-19 and for the war in Ukraine, two topics for which the government's communication is prevalent.
Corruption is one notable exception, with the center-right being most cited because several of its politicians, such as the three with the lowest average scores from Table~\ref{tab:low_high} were involved in major corruption scandals.
There are some differences even for closely related topics, such as the pairs related to Covid-19 and Ukraine. 
There, the mentions of the center are more dominant for Covid Health and for Ukraine War.

The deviation of sentiment associated with political orientations from the average of the topic, illustrated in Figure~\ref{fig:topics_sentiment}, can be interpreted as a proxy for the credibility of political orientations for the respective topic.
For instance, corruption has a particularly negative representation, with an average sentiment of -0.5. 
While still in the negative range, RL, CL and C orientations are perceived better than CR and RR on this crucial political topic. 
The radical right has the lowest scores on average, with particularly negative positioning for climate change and the war in Ukraine.
This finding is probably explained by the relatively small importance given to environment in RR platforms and by the favorable positioning with respect to Russia in the past for the war in Ukraine. 
The radical left is also perceived negatively on the two Ukraine related topics for a similar reason.
However, the sentiment toward RL is more positive than that of other orientations for most other topics.

\subsection{Politician Oriented Analysis} 
\label{subsec:person}

\noindent\textbf{Frequently-mentioned Politicians.}
Table~\ref{tab:top_10_politicians} provides insights about the top-10 most frequently mentioned politicians: overall mentions, average sentiment in the corpus, sentiment variation over time (i.e. the standard deviation of sentiment per year between 2016 and 2022). 
The table shows that political discussion in France is focused on presidency since all top-10 names are of French or foreign presidents or presidential candidates. 
To verify the potential bias due to presidential elections, we ran the analysis separately for years without election and results are similar.
\textit{Emmanuel Macron} is by far the most mentioned politician, with Russian and American presidents also being mentioned frequently.
The other French politicians from Table~\ref{tab:top_10_politicians} are mainly 2022 presidential candidates.
There is no correlation between their mentions in the corpus and their electoral results. 
\textit{Eric Zemmour} and \textit{Valérie Pécresse} scored lower in the elections than \textit{Marine Le Pen} and \textit{Jean-Luc Mélenchon}, but are more frequently mentioned. 

\begin{table}
	\begin{center}
		\resizebox{0.99\linewidth}{!}{
\begin{tabular}{|l|c|c|c|}
\hline
\multirow{2}{*}{\textbf{Name}}& \multicolumn{2}{|c|}{Entity statistics} & Sentiment  \\ \cline{2-3}
 & Mentions & Sentiment & variation \\ \hline
Emmanuel Macron & 49552 & -0.074 & 0.036 \\ \hline
Vladimir Putin & 22925 & -0.355 & 0.085 \\  \hline
Éric Zemmour & 21201 & -0.176 & 0.106 \\ \hline
Valérie Pécresse & 15055 & -0.006 & 0.09 \\ \hline
Jean-Luc Mélenchon & 12223 & -0.052  & 0.097 \\ \hline
Joe Biden & 9907 & -0.035 & 0.159   \\ \hline
Donald Trump & 8383 & -0.430 & 0.071 \\ \hline
Anne Hidalgo & 7539 & -0.090 & 0.032 \\ \hline
Nicolas Sarkozy & 6852 & -0.319 & 0.042  \\ \hline
Marine Le Pen & 6639 & -0.265 & 0.066 \\ \hline
\end{tabular}			
		}
	\end{center}
	\caption{Top-10 most frequently politicians, with: number of mentions; average sentiment scores computed using the entire corpus; standard deviation of average sentiment per year between 2016 and 2022.}
	\label{tab:top_10_politicians}
\end{table} 

The average sentiment scores from Table~\ref{tab:top_10_politicians} vary substantially. 
\textit{Donald Trump}, \textit{Vladimir Putin} and \textit{Nicolas Sarkozy} have the lowest scores.
This is probably due to the critical perception of their political actions for the first two, and by ongoing legal actions against the third. 
Interestingly, \textit{Valérie Pécresse} has the highest sentiment score in Table~\ref{tab:top_10_politicians}, despite her poor electoral performance in 2022. 
We verified her mentions and many of them correspond to her unexpected victory in the primary election of her party. 
We note that sentiment is more negative for far-right candidates (\textit{Eric Zemmour} and \textit{Marine Le Pen}) compared to \textit{Jean-Luc Mélenchon}, the main left-wing candidate. 
This finding confirms the aggregated results from Figure~\ref{fig:mention_positioning}.
Sentiment varies to a different extent during the examined period, and this is explained by different factors: the political comeback for \textit{Joe Biden}, the emergence of \textit{Eric Zemmour} as presidential candidate in 2022, the invasion of Ukraine for \textit{Vladimir Putin} etc.
Details about temporal dynamics of mentions and sentiment for politicians from Table~\ref{tab:top_10_politicians} are provided in Appendix~\ref{app:temporal_politicians}.

\begin{table*}[ht]
	\begin{center}
			\begin{tabular}{|p{0.1\textwidth}|p{0.8\textwidth}|}
\hline
\textbf{Scores} & \textbf{Politician name (average score)} \\ \hline
Highest & \textcolor{teal}{Valéry Giscard (0.239); Carole Delga (0.193); Jean Lassalle (0.180); Roselyne Bachelot (0.163); Olaf Scholz (0.139); David Lisnard (0.135); Kamala Harris (0.125); Jerome Powell (0.104); Julien Denormandie (0.096); Michel Barnier (0.093)} \\ \hline
Lowest &  \textcolor{red}{Claude Guéant (-0.70); Penelope Fillon (-0.61); Patrick Balkany (-0.52); Alexander Lukashenko (-0.50); Bernard Tapie (-0.48); Donald Trump (-0.43); Nicolas Hulot (-0.40); Jair Bolsonaro (-0.39); Vladimir Putin (-0.35); Viktor Orbán (-0.34)} \\ \hline
\end{tabular}
	\end{center}
	\caption{Politicians with the lowest and highest average sentiment scores (bottom and top of the figure, respectively). They were selected from the subset of the top-100 most frequently mentions politicians in our corpus.}
	\label{tab:low_high}
\vspace{-5mm} 
\end{table*} 

\noindent\textbf{Polarized Sentiment Scores.}
Table~\ref{tab:low_high} presents the top-10 politicians with the highest and lowest sentiment scores. 
The first group includes local politicians whose action is well-appreciated (\textit{Carole Delga} and \textit{David Lisnard}), national politicians who are favorably viewed by the public (\textit{Roselyne Bachelot} and \textit{Jean Lassalle}), and international politicians with a centrist or center-left positioning who took office during the analyzed period (\textit{Olaf Scholtz} and \textit{Kamala Harris}). 
\textit{Valéry Giscard} is an exception in that this former French president died in 2020 and was already retired from politics for a long time. 
These two factors could explain his favorable representation in the corpus. 
Strongly negative scores are mainly associated with politicians involved in prominent political scandals (\textit{Claude Guéant, Penelope Fillon, Patrick Balkany}) and to authoritarian leaders (\textit{Alexander Lukashenko}) or populist leaders (\textit{Donald Trump, Jair Bolsonaro, Viktor Orban}). 
We note that \textit{Penelope Fillon} has a much more negative score (-0.61) than \textit{François Fillon} (-0.33), her husband, who provided her a no-show job for decades. 
This is explained by the fact his representation in the corpus includes the beginning of the 2017 presidential campaign, when \textit{François Fillon} won his party's primary elections.

\subsection{Gender Representation}
\label{subsec:gender}
\begin{figure}[htbp]
	\centering
	\includegraphics[width=0.99\linewidth, trim={0cm 0cm 0cm 0cm}]{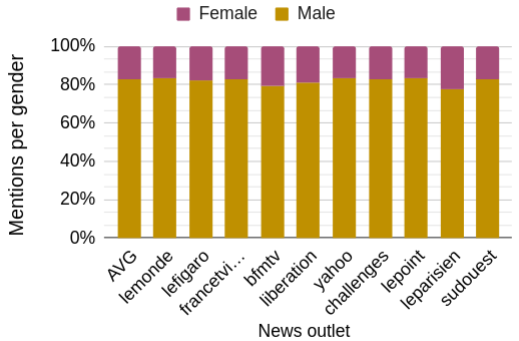} 
 \vspace{-4mm}
	\caption{Percentage of gender mentions in the top-10 news outlets and on average.
	}
	\label{fig:gender_mentions}
\end{figure}

\begin{figure}[htb]
	\centering
	\includegraphics[width=0.99\linewidth, trim={0cm 0cm 0cm 0cm}]{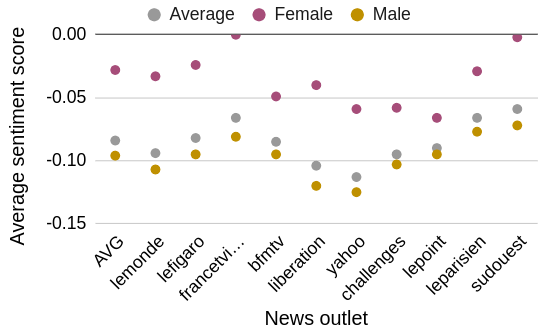} 
 \vspace{-4mm}
	\caption{Mean sentiment classification scores by gender in the top-10 news outlets and on average.}
	\label{fig:gender_sentiment}
\end{figure}

Existing studies which quantify the representation of genders in politics~\cite{asr2021gender,doukhan2018describing} show that men are much more present than women. 
We deepen this analysis by adding a subjective component to understand how female and male politicians are represented in the news~\footnote{The corpus includes only 8 mentions of other gender.}, with an illustration of mentions (Figure~\ref{fig:gender_mentions}) and mean sentiment (Figure ~\ref{fig:gender_sentiment}).
The results from Figure~\ref{fig:gender_mentions} show that the news from the analyzed corpus contain significantly more mentions of male than of female politicians. 
The percentage of female mentions varies from 16.7\% (\textit{lemonde}) to 22.3\% (\textit{leparisien}). 
Considering that the overall proportion of women in French politics is higher (38\% in Parliament\footnote{\url{https://data.ipu.org/content/france}} and 50\% in the Government), the results from Figure~\ref{fig:gender_mentions} show a clear underrepresentation of women.
This bias is higher than the one observed for French audio-visual media, where female politicians amount for 30\% of the total mentions~\cite{arcom_2021}.
The difference is probably explained by the stronger regulatory pressure on audio-visual media compared to written media. 
A temporal analysis of gender representation between 2016 and 2022 (detailed in Appendix~\ref{app:temporal_gender}) shows that gender bias is relatively stable.
This finding indicates that the debate about the representation of gender in society have only a limited effect on the quantitative representation of women in politics. 

The gender-focused analysis of sentiment from Figure~\ref{fig:gender_sentiment} shows that the mentions of female politicians are more favorable than those of their male counterparts. 
This trend is respected for individual outlets, with some variation of the gap between genders.
This favorable coverage is a positive signal for reducing gender bias in politics, but more efforts are needed to reduce the mention gap. 

\begin{figure}[htb]
	\centering	\includegraphics[width=0.99\linewidth, trim={0cm 0cm 0cm 0cm}]{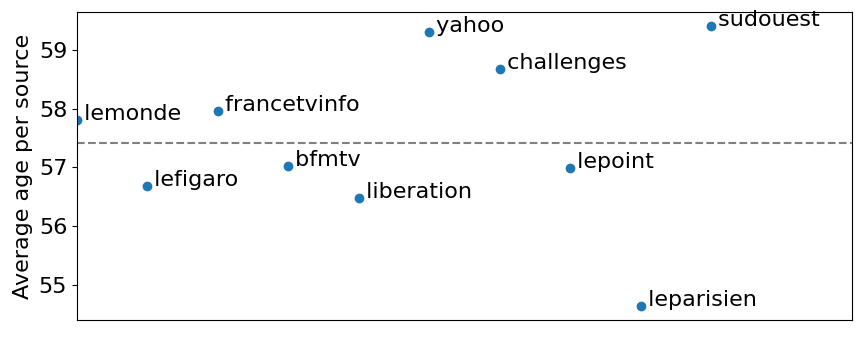} 
	\caption{Average age of the politicians mentioned at the time of publication per news outlet.
	}
	\label{fig:age_mentions}
\end{figure}
\subsection{Age Representation}
Figure~\ref{fig:age_mentions} illustrates the average age of politicians mentioned in the corpus. 
The global average is 57.4 years, with individual averages varying between 54.6 for \textit{leparisien} and 59.4 for \textit{sudouest}. 
The average age of members of the French Parliament is 49.1, while the corresponding figure is even lower for the Government.
This age-related bias confirms the qualitative results presented in Table~\ref{tab:top_10_politicians}, which shows that the majority of frequently mentioned politicians are older than the average.
This is particularly the case for \textit{Joe Biden, Donald Trump, Vladimir Putin} and \textit{Nicolas Sarkozy}.
The main exception is \textit{Emmanuel Macron}, who is much younger than the average age reported in Figure~\ref{fig:age_mentions}.

\section{Limitations}
\label{sec:limitations}
We discuss below a series of limitations of the proposed analysis framework, both in terms of data and processing.

\noindent\textbf{Scope of the dataset.} 
The collected dataset includes a diversity of French media which are available online, but is focused on classical media.
It does not include political texts published via social networks, which could further increase diversity.
Such an enrichment of the dataset is challenging due to the content access policies of these platforms. 
The dataset content might induce biases regarding the results of the analysis. However, we carefully designed experiments so as to have sufficiently large samples in each case.

\noindent\textbf{Dataset imbalance.} 
The sources are represented by a variable number of news and of mentions of entities, as detailed in Appendix~\ref{app:statistics}. 
This is an effect of the collection strategy, which was more or less intensive for different time intervals, of the availability of RSS streams, and of the existence of paywalls.
In the latter case, only the first sentences of the articles are available, but this block of text still provides interesting information. 
The outlet-related imbalance is mitigated by the fact that the analyses are performed on aggregated samples of data with sufficient data point for each sample.

\noindent\textbf{Types of entities used for analysis.}
The entity detection and linking components handle persons, organizations, and locations.
The first two types of entities are most interesting for our work, but the sentiment classification dataset only includes persons. 
The extension to organizations, political parties in particular, would be useful.
However, it would require an important labeling effort and is left for future work. 
The absence of organizations is mitigated by the fact that politicians can be mapped to political parties using Wikidata~\cite{pellissier2016freebase}, which can be themselves assigned to political orientations via ParlGov~\cite{doring2012parliament}.

\noindent\textbf{Sentiment classification robustness.}
In Section~\ref{section:pipeline}, we show that sentiment classification leads to stable aggregated results, despite imperfections for individual cases.
Another potential limitation is related to the robustness of results when inferring sentiment for a set of entities that are not in the training set.
This is the case for the news corpus used here since it is independent of the sentiment classification dataset used for training this component.
We perform a train/validation/test split of the target-dependent classification dataset in which the validation and test subsets contained 50\% of entities that appear in the training set and 50\% which do not appear in it. 
After inference, we compute the $F1$ scores separately for the two test subsets and the difference in performance is under 1\%.
We also manually annotate a set of 500 randomly-selected sentiment predictions from our corpus and again obtained $F1$ scores which are similar to those obtained for the internal test set from~\citet{dufraisseMADTSCMultilingualAligned2023}.
This indicates that sentiment classification is transferable to new entities and can be used here.

\noindent\textbf{Completeness of analysis.}
The dimensions of the news representations obtained with the our framework can be combined in a flexible way.
A series of results were presented in Section~\ref{sec:analysis}, and they are complemented by those discussed in the Appendices. 
Other combinations of dimensions of news representations could be used to perform different analyses, for instance by aggregating results at the political party level, but cannot be presented due to lack of space.
We will release the code and data needed to facilitate further experiments by experts from different disciplines which take an interest in political news analysis. 

\section{Conclusions}
We introduced a news analysis framework which is useful for different computational politics stakeholders.
The proposed framework combines recent NLP components, information retrieval and structured information to provide rich insights about the representation of politics in the news. 
The NLP components are multilingual and the resources ensure international coverage. 
The approach could thus be easily transferred to other countries and other languages in order to understand the similarities and differences of political representations in different democratic countries.
We provide a comprehensive analysis focused on sources, politicians, and the representation of demographic segments. 
The obtained results are coherent with past findings~\cite{hosting_cag_2022,doukhan2018describing}, but are significantly different, notably due to the important role given to the subjective perspective enabled by the use of target-dependent sentiment classification. 
At source level, we quantify the sentiment expressed of outlets and unveil their positioning with respect to the main political orientations.
At topic level, we show that there are important differences between the distributions of mentions of political orientations and that the sentiment expressed about them varies significantly.
At an individual level, we confirm that the French political representation in the media is dominated by the presidential function, and show that the automatic analysis is well-correlated with the public perception of politicians.
Regarding demographic biases, we note that, unfortunately, the representation of politics in the French news is still a ``country for old men'', since young and female politicians are still strongly underrepresented. 

These findings are only examples of the insights which can be obtained with a rich and flexible representation of news, and can be used by multiple stakeholders.
Newspaper editors and journalists can obtain feedback about their outlet's positioning and that of competitors, and push toward a more diversified coverage of political news.
Social scientists can use this type of insights to deepen the analysis of the presentation of politicians and politics in the media.
Political organizations can deploy the pipeline to monitor their political action and improve it.
Regulatory bodies could use framework to quantify biases and push toward a more diversified representation of demographic segments. 
The proposed analyses are also a way to improve transparency about the political topics and positioning of news outlets. 
As such, they can improve the understanding of the political news landscape by the general public. 

Code and data used here will be released to enable reproducibility. 
The links of news included in the corpus will also be provided.


\bibliographystyle{ACM-Reference-Format}
\bibliography{custom,anthology,sample-base,aaai22}

\begin{thebibliography}{60}
\providecommand{\natexlab}[1]{#1}

\bibitem[{Agrawal et~al.(2022)Agrawal, Gupta, Gautam, and Mamidi}]{agrawal-etal-2022-towards}
Agrawal, S.; Gupta, K.; Gautam, D.; and Mamidi, R. 2022.
\newblock Towards Detecting Political Bias in {H}indi News Articles.
\newblock In \emph{Proceedings of the 60th Annual Meeting of the Association for Computational Linguistics: Student Research Workshop}, 239--244. Dublin, Ireland: Association for Computational Linguistics.

\bibitem[{Akbik et~al.(2019)Akbik, Bergmann, Blythe, Rasul, Schweter, and Vollgraf}]{akbik-etal-2019-flair}
Akbik, A.; Bergmann, T.; Blythe, D.; Rasul, K.; Schweter, S.; and Vollgraf, R. 2019.
\newblock {FLAIR}: An Easy-to-Use Framework for State-of-the-Art {NLP}.
\newblock In \emph{Proceedings of the 2019 Conference of the North {A}merican Chapter of the Association for Computational Linguistics (Demonstrations)}, 54--59. Minneapolis, Minnesota: Association for Computational Linguistics.

\bibitem[{Alonso~del Barrio and Gatica-Perez(2023)}]{alonso2023framing}
Alonso~del Barrio, D.; and Gatica-Perez, D. 2023.
\newblock Framing the News: From Human Perception to Large Language Model Inferences.
\newblock In \emph{Proceedings of the 2023 ACM International Conference on Multimedia Retrieval}, 627--635.

\bibitem[{ARCOM(2021)}]{arcom_2021}
ARCOM. 2021.
\newblock The representation of women on French TV and radio stations - 2021 report (in French).
\newblock \url{https://www.arcom.fr/sites/default/files/2022-03/La\%20repr\%C3\%A9sentation\%20des\%20femmes\%20\%C3\%A0\%20la\%20t\%C3\%A9l\%C3\%A9vision\%20et\%20\%C3\%A0\%20la\%20radio\%20-\%20Rapport\%20sur\%20l\%27exercice\%202021_0.pdf}.
\newblock Accessed: 2023-06-10.

\bibitem[{Asr et~al.(2021)Asr, Mazraeh, Lopes, Gautam, Gonzales, Rao, and Taboada}]{asr2021gender}
Asr, F.~T.; Mazraeh, M.; Lopes, A.; Gautam, V.; Gonzales, J.; Rao, P.; and Taboada, M. 2021.
\newblock The gender gap tracker: Using natural language processing to measure gender bias in media.
\newblock \emph{PloS one}, 16(1): e0245533.

\bibitem[{Balahur et~al.(2010)Balahur, Steinberger, Kabadjov, Zavarella, van~der Goot, Halkia, Pouliquen, and Belyaeva}]{balahur-etal-2010-sentiment}
Balahur, A.; Steinberger, R.; Kabadjov, M.; Zavarella, V.; van~der Goot, E.; Halkia, M.; Pouliquen, B.; and Belyaeva, J. 2010.
\newblock Sentiment Analysis in the News.
\newblock In \emph{Proceedings of the Seventh International Conference on Language Resources and Evaluation ({LREC}'10)}. Valletta, Malta: European Language Resources Association (ELRA).

\bibitem[{Baly et~al.(2020)Baly, Martino, Glass, and Nakov}]{baly2020we}
Baly, R.; Martino, G. D.~S.; Glass, J.; and Nakov, P. 2020.
\newblock We can detect your bias: Predicting the political ideology of news articles.
\newblock \emph{arXiv preprint arXiv:2010.05338}.

\bibitem[{Barbaresi(2021)}]{barbaresi-2021-trafilatura}
Barbaresi, A. 2021.
\newblock Trafilatura: {A} Web Scraping Library and Command-Line Tool for Text Discovery and Extraction.
\newblock In \emph{Proceedings of the 59th Annual Meeting of the Association for Computational Linguistics and the 11th International Joint Conference on Natural Language Processing: System Demonstrations}, 122--131. Online: Association for Computational Linguistics.

\bibitem[{Bastin(2022)}]{bastin_gender_2022}
Bastin, G. 2022.
\newblock Gender Imbalance in the Media: Time Lag or Hysteresis?—French Newspapers, Gender Parity Shocks, and the Long and Winding Road to the Demasculinization of Political Reporting (1990–2020).
\newblock \emph{The International Journal of Press/Politics}, 0(0): 19401612221143074.

\bibitem[{Bautin, Vijayarenu, and Skiena()}]{bautinInternationalSentimentAnalysis2021}
Bautin, M.; Vijayarenu, L.; and Skiena, S. ????
\newblock International {{Sentiment Analysis}} for {{News}} and {{Blogs}}.
\newblock 2(1): 19--26.

\bibitem[{Bestvater and Monroe(2023)}]{bestvater_monroe_2023}
Bestvater, S.~E.; and Monroe, B.~L. 2023.
\newblock Sentiment is Not Stance: Target-Aware Opinion Classification for Political Text Analysis.
\newblock \emph{Political Analysis}, 31(2): 235--256.

\bibitem[{Brauwers and Frasincar(2022)}]{brauwers2022survey}
Brauwers, G.; and Frasincar, F. 2022.
\newblock A Survey on Aspect-Based Sentiment Classification.
\newblock \emph{ACM Comput. Surv.}, 55(4).

\bibitem[{Broder(1998)}]{broder1998}
Broder, A. 1998.
\newblock On the Resemblance and Containment of Documents.
\newblock In \emph{Proceedings. {{Compression}} and {{Complexity}} of {{SEQUENCES}} 1997 ({{Cat}}. {{No}}.{{97TB100171}})}, 21--29. Salerno, Italy: IEEE Comput. Soc.
\newblock ISBN 978-0-8186-8132-5.

\bibitem[{Cag\'{e} et~al.(2022)Cag\'{e}, Hengel, Herv\'{e}, and Urvoy}]{hosting_cag_2022}
Cag\'{e}, J.; Hengel, M.; Herv\'{e}, N.; and Urvoy, C. 2022.
\newblock Hosting Media Bias: Evidence from the Universe of French Broadcasts, 2002-2020.
\newblock \emph{SSRN Electronic Journal}.

\bibitem[{Cointet et~al.(2021)Cointet, Cardon, Mogoutov, Ooghe-Tabanou, Plique, and Morales}]{cointet2021uncovering}
Cointet, J.-P.; Cardon, D.; Mogoutov, A.; Ooghe-Tabanou, B.; Plique, G.; and Morales, P. 2021.
\newblock Uncovering the structure of the French media ecosystem.
\newblock \emph{arXiv preprint arXiv:2107.12073}.

\bibitem[{Conforti et~al.(2020)Conforti, Berndt, Pilehvar, Giannitsarou, Toxvaerd, and Collier}]{conforti-etal-2020-stander}
Conforti, C.; Berndt, J.; Pilehvar, M.~T.; Giannitsarou, C.; Toxvaerd, F.; and Collier, N. 2020.
\newblock {STANDER}: An Expert-Annotated Dataset for News Stance Detection and Evidence Retrieval.
\newblock In \emph{Findings of the Association for Computational Linguistics: EMNLP 2020}, 4086--4101. Online: Association for Computational Linguistics.

\bibitem[{Dacon and Liu(2021)}]{dacon2021}
Dacon, J.; and Liu, H. 2021.
\newblock Does Gender Matter in the News? Detecting and Examining Gender Bias in News Articles.
\newblock In \emph{Companion Proceedings of the Web Conference 2021}, WWW '21, 385--392. Ljubljana, Slovenia: Association for Computing Machinery.
\newblock ISBN 9781450383134.

\bibitem[{De~Cao et~al.(2022)De~Cao, Wu, Popat, Artetxe, Goyal, Plekhanov, Zettlemoyer, Cancedda, Riedel, and Petroni}]{de-cao-etal-2022-multilingual}
De~Cao, N.; Wu, L.; Popat, K.; Artetxe, M.; Goyal, N.; Plekhanov, M.; Zettlemoyer, L.; Cancedda, N.; Riedel, S.; and Petroni, F. 2022.
\newblock Multilingual Autoregressive Entity Linking.
\newblock \emph{Transactions of the Association for Computational Linguistics}, 10: 274--290.

\bibitem[{Deshmukh and Sethi(2020)}]{deshmukh2020}
Deshmukh, A.~A.; and Sethi, U. 2020.
\newblock {{IR-BERT}}: {{Leveraging BERT}} for {{Semantic Search}} in {{Background Linking}} for {{News Articles}}.

\bibitem[{Devlin et~al.(2019)Devlin, Chang, Lee, and Toutanova}]{devlin-etal-2019-bert}
Devlin, J.; Chang, M.-W.; Lee, K.; and Toutanova, K. 2019.
\newblock {BERT}: Pre-training of Deep Bidirectional Transformers for Language Understanding.
\newblock In \emph{Proceedings of the 2019 Conference of the North {A}merican Chapter of the Association for Computational Linguistics: Human Language Technologies, Volume 1 (Long and Short Papers)}, 4171--4186. Minneapolis, Minnesota: Association for Computational Linguistics.

\bibitem[{D\"{o}ring and Manow(2012)}]{doring2012parliament}
D\"{o}ring, H.; and Manow, P. 2012.
\newblock Parliament and government composition database (ParlGov).
\newblock \emph{An infrastructure for empirical information on parties, elections and governments in modern democracies. Version}, 12(10).

\bibitem[{Doukhan et~al.(2018)Doukhan, Poels, Rezgui, and Carrive}]{doukhan2018describing}
Doukhan, D.; Poels, G.; Rezgui, Z.; and Carrive, J. 2018.
\newblock Describing gender equality in french audiovisual streams with a deep learning approach.
\newblock \emph{VIEW Journal of European Television History and Culture}, 7(14): 103--122.

\bibitem[{Dufraisse et~al.()Dufraisse, Popescu, Tourille, Brun, and Deshayes}]{dufraisseMADTSCMultilingualAligned2023}
Dufraisse, E.; Popescu, A.; Tourille, J.; Brun, A.; and Deshayes, J. ????
\newblock {{MAD-TSC}}: {{A Multilingual Aligned News Dataset}} for {{Target-dependent Sentiment Classification}}.
\newblock In \emph{Proceedings of the 61st {{Annual Meeting}} of the {{Association}} for {{Computational Linguistics}} ({{Volume}} 1: {{Long Papers}})}, 8286--8305. {Association for Computational Linguistics}.

\bibitem[{Entman(1993)}]{entmanFramingClarificationFractured1993}
Entman, R.~M. 1993.
\newblock {Framing: Toward Clarification of a Fractured Paradigm}.
\newblock \emph{Journal of Communication}, 43(4): 51--58.

\bibitem[{{FORCE11}(2020)}]{fair}
{FORCE11}. 2020.
\newblock The FAIR Data principles.
\newblock \url{https://force11.org/info/the-fair-data-principles/}.

\bibitem[{Gangula, Duggenpudi, and Mamidi(2019)}]{gangula-etal-2019-detecting}
Gangula, R. R.~R.; Duggenpudi, S.~R.; and Mamidi, R. 2019.
\newblock Detecting Political Bias in News Articles Using Headline Attention.
\newblock In \emph{Proceedings of the 2019 ACL Workshop BlackboxNLP: Analyzing and Interpreting Neural Networks for NLP}, 77--84. Florence, Italy: Association for Computational Linguistics.

\bibitem[{Gebru et~al.(2021)Gebru, Morgenstern, Vecchione, Vaughan, Wallach, Iii, and Crawford}]{gebru2021datasheets}
Gebru, T.; Morgenstern, J.; Vecchione, B.; Vaughan, J.~W.; Wallach, H.; Iii, H.~D.; and Crawford, K. 2021.
\newblock Datasheets for datasets.
\newblock \emph{Communications of the ACM}, 64(12): 86--92.

\bibitem[{Godbole, Srinivasaiah, and Skiena()}]{godboleLargeScaleSentimentAnalysis}
Godbole, N.; Srinivasaiah, M.; and Skiena, S. ????
\newblock Large-{{Scale Sentiment Analysis}} for {{News}} and {{Blogs}} ({{System Demonstration}}).

\bibitem[{Hamborg(2020)}]{hamborg-2020-media}
Hamborg, F. 2020.
\newblock Media Bias, the Social Sciences, and {NLP}: Automating Frame Analyses to Identify Bias by Word Choice and Labeling.
\newblock In \emph{Proceedings of the 58th Annual Meeting of the Association for Computational Linguistics: Student Research Workshop}, 79--87. Online: Association for Computational Linguistics.

\bibitem[{Hamborg(2023)}]{Hamborg2023}
Hamborg, F. 2023.
\newblock \emph{Person-Oriented Framing Analysis}, 55--70.
\newblock Cham: Springer Nature Switzerland.
\newblock ISBN 978-3-031-17693-7.

\bibitem[{Hamborg and Donnay(2021)}]{hamborg-donnay-2021-newsmtsc}
Hamborg, F.; and Donnay, K. 2021.
\newblock {N}ews{MTSC}: A Dataset for (Multi-)Target-dependent Sentiment Classification in Political News Articles.
\newblock In \emph{Proceedings of the 16th Conference of the European Chapter of the Association for Computational Linguistics: Main Volume}, 1663--1675. Online: Association for Computational Linguistics.

\bibitem[{Hamborg, Donnay, and Gipp(2019)}]{hamborg2019automated}
Hamborg, F.; Donnay, K.; and Gipp, B. 2019.
\newblock Automated identification of media bias in news articles: an interdisciplinary literature review.
\newblock \emph{International Journal on Digital Libraries}, 20(4): 391--415.

\bibitem[{Kamal et~al.(2023)Kamal, Hartford, Willis, and Bagavathi}]{kamal2023learning}
Kamal, S.; Hartford, J.; Willis, J.; and Bagavathi, A. 2023.
\newblock Learning Unbiased News Article Representations: A Knowledge-Infused Approach.
\newblock \emph{arXiv preprint arXiv:2309.05981}.

\bibitem[{Khloponin and Kosseim(2021)}]{khloponin2021}
Khloponin, P.; and Kosseim, L. 2021.
\newblock Using {{Document Embeddings}} for {{Background Linking}} of {{News Articles}}.
\newblock 12801: 317--329.

\bibitem[{Kim, Lelkes, and McCrain(2022)}]{measuring_kim_2022}
Kim, E.; Lelkes, Y.; and McCrain, J. 2022.
\newblock Measuring dynamic media bias.
\newblock \emph{Proceedings of the National Academy of Sciences of the United States of America}.

\bibitem[{Ko et~al.(2023)Ko, Ryu, Han, Jeon, Kim, Park, Han, Tong, and Kim}]{ko2023khan}
Ko, Y.; Ryu, S.; Han, S.; Jeon, Y.; Kim, J.; Park, S.; Han, K.; Tong, H.; and Kim, S.-W. 2023.
\newblock KHAN: Knowledge-Aware Hierarchical Attention Networks for Political Stance Prediction.
\newblock \emph{arXiv preprint arXiv:2302.12126}.

\bibitem[{K{\"u}{\c{c}}{\"u}k and Can(2020)}]{kuccuk2020stance}
K{\"u}{\c{c}}{\"u}k, D.; and Can, F. 2020.
\newblock Stance detection: A survey.
\newblock \emph{ACM Computing Surveys (CSUR)}, 53(1): 1--37.

\bibitem[{Lenoir(2019)}]{lenoir2019montaigne}
Lenoir, T. 2019.
\newblock Media polarization "à la française" ? Comparing the French and American ecosystems.
\newblock Technical report, Institut Montaigne.

\bibitem[{Liu(2010)}]{liu2010sentiment}
Liu, B. 2010.
\newblock Sentiment analysis and subjectivity.
\newblock \emph{Handbook of natural language processing}, 2(2010): 627--666.

\bibitem[{Liu et~al.(2019{\natexlab{a}})Liu, Guo, Mays, Betke, and Wijaya}]{liu-etal-2019-detecting}
Liu, S.; Guo, L.; Mays, K.; Betke, M.; and Wijaya, D.~T. 2019{\natexlab{a}}.
\newblock Detecting Frames in News Headlines and Its Application to Analyzing News Framing Trends Surrounding {U}.{S}. Gun Violence.
\newblock In \emph{Proceedings of the 23rd Conference on Computational Natural Language Learning (CoNLL)}, 504--514. Hong Kong, China: Association for Computational Linguistics.

\bibitem[{Liu et~al.(2019{\natexlab{b}})Liu, Ott, Goyal, Du, Joshi, Chen, Levy, Lewis, Zettlemoyer, and Stoyanov}]{liu2019roberta}
Liu, Y.; Ott, M.; Goyal, N.; Du, J.; Joshi, M.; Chen, D.; Levy, O.; Lewis, M.; Zettlemoyer, L.; and Stoyanov, V. 2019{\natexlab{b}}.
\newblock Roberta: A robustly optimized bert pretraining approach.
\newblock \emph{arXiv preprint arXiv:1907.11692}.

\bibitem[{Lloyd, Kechagias, and Skiena(2005)}]{lloyd2005lydia}
Lloyd, L.; Kechagias, D.; and Skiena, S. 2005.
\newblock Lydia: A system for large-scale news analysis.
\newblock In \emph{International Symposium on String Processing and Information Retrieval}, 161--166. Springer.

\bibitem[{Martin and Yurukoglu(2017)}]{10.1257/aer.20160812}
Martin, G.~J.; and Yurukoglu, A. 2017.
\newblock Bias in Cable News: Persuasion and Polarization.
\newblock \emph{American Economic Review}, 107(9): 2565--99.

\bibitem[{Martin et~al.(2020)Martin, Muller, Ortiz~Su{\'a}rez, Dupont, Romary, de~la Clergerie, Seddah, and Sagot}]{martin-etal-2020-camembert}
Martin, L.; Muller, B.; Ortiz~Su{\'a}rez, P.~J.; Dupont, Y.; Romary, L.; de~la Clergerie, {\'E}.; Seddah, D.; and Sagot, B. 2020.
\newblock {C}amem{BERT}: a Tasty {F}rench Language Model.
\newblock In \emph{Proceedings of the 58th Annual Meeting of the Association for Computational Linguistics}, 7203--7219. Online: Association for Computational Linguistics.

\bibitem[{Nakov et~al.(2016)Nakov, Ritter, Rosenthal, Sebastiani, and Stoyanov}]{nakov-etal-2016-semeval}
Nakov, P.; Ritter, A.; Rosenthal, S.; Sebastiani, F.; and Stoyanov, V. 2016.
\newblock {S}em{E}val-2016 Task 4: Sentiment Analysis in {T}witter.
\newblock In \emph{Proceedings of the 10th International Workshop on Semantic Evaluation ({S}em{E}val-2016)}, 1--18. San Diego, California: Association for Computational Linguistics.

\bibitem[{Pellissier~Tanon et~al.(2016)Pellissier~Tanon, Vrande\v{c}i\'{c}, Schaffert, Steiner, and Pintscher}]{pellissier2016freebase}
Pellissier~Tanon, T.; Vrande\v{c}i\'{c}, D.; Schaffert, S.; Steiner, T.; and Pintscher, L. 2016.
\newblock From freebase to wikidata: The great migration.
\newblock In \emph{Proceedings of the 25th international conference on world wide web}, 1419--1428.

\bibitem[{Ploch, Lommatzsch, and Schultze()}]{plochAdvancedPressReview2016}
Ploch, D.; Lommatzsch, A.; and Schultze, F. ????
\newblock An {{Advanced Press Review System Combining Deep News Analysis}} and {{Machine Learning Algorithms}}.
\newblock In \emph{Proceedings of {{ACL-2016 System Demonstrations}}}, 109--114. {Association for Computational Linguistics}.

\bibitem[{Pontiki et~al.(2016)Pontiki, Galanis, Papageorgiou, Androutsopoulos, Manandhar, AL-Smadi, Al-Ayyoub, Zhao, Qin, De~Clercq, Hoste, Apidianaki, Tannier, Loukachevitch, Kotelnikov, Bel, Jim{\'e}nez-Zafra, and Eryi{\u{g}}it}]{pontiki-etal-2016-semeval}
Pontiki, M.; Galanis, D.; Papageorgiou, H.; Androutsopoulos, I.; Manandhar, S.; AL-Smadi, M.; Al-Ayyoub, M.; Zhao, Y.; Qin, B.; De~Clercq, O.; Hoste, V.; Apidianaki, M.; Tannier, X.; Loukachevitch, N.; Kotelnikov, E.; Bel, N.; Jim{\'e}nez-Zafra, S.~M.; and Eryi{\u{g}}it, G. 2016.
\newblock {S}em{E}val-2016 Task 5: Aspect Based Sentiment Analysis.
\newblock In \emph{Proceedings of the 10th International Workshop on Semantic Evaluation ({S}em{E}val-2016)}, 19--30. San Diego, California: Association for Computational Linguistics.

\bibitem[{Richard, Bastin, and Portet(2022)}]{richard2022genderednews}
Richard, A.; Bastin, G.; and Portet, F. 2022.
\newblock GenderedNews: Une approche computationnelle des \'ecarts de repr\'esentation des genres dans la presse fran\c{c}aise.
\newblock arXiv:2202.05682.

\bibitem[{Rivas-de Roca, Caro-Gonz\'{a}lez, and Garc\'{\i}a-Gordillo(2022)}]{rivas_eu}
Rivas-de Roca, R.; Caro-Gonz\'{a}lez, F.~J.; and Garc\'{\i}a-Gordillo, M. 2022.
\newblock Covering the EU at local level: A multiple-case study in Germany, the UK and Spain.
\newblock \emph{International Communication Gazette}, 17480485221134177.

\bibitem[{Robertson, Zaragoza et~al.(2009)}]{robertson2009probabilistic}
Robertson, S.; Zaragoza, H.; et~al. 2009.
\newblock The probabilistic relevance framework: BM25 and beyond.
\newblock \emph{Foundations and Trends{\textregistered{}{}{}} in Information Retrieval}, 3(4): 333--389.

\bibitem[{Sep\'{u}lveda-Torres et~al.(2023)Sep\'{u}lveda-Torres, Vicente, Saquete, Lloret, and Palomar}]{torres_stance}
Sep\'{u}lveda-Torres, R.; Vicente, M.; Saquete, E.; Lloret, E.; and Palomar, M. 2023.
\newblock Leveraging relevant summarized information and multi-layer classification to generalize the detection of misleading headlines.
\newblock \emph{Data \& Knowledge Engineering}, 145: 102176.

\bibitem[{Severyn et~al.(2016)Severyn, Moschitti, Uryupina, Plank, and Filippova}]{severyn2016multi}
Severyn, A.; Moschitti, A.; Uryupina, O.; Plank, B.; and Filippova, K. 2016.
\newblock Multi-lingual opinion mining on YouTube.
\newblock \emph{Information Processing \& Management}, 52(1): 46--60.

\bibitem[{Song et~al.(2019)Song, Wang, Jiang, Liu, and Rao}]{song2019attentional}
Song, Y.; Wang, J.; Jiang, T.; Liu, Z.; and Rao, Y. 2019.
\newblock Attentional encoder network for targeted sentiment classification.
\newblock \emph{arXiv preprint arXiv:1902.09314}.

\bibitem[{Spinde, Hamborg, and Gipp(2020)}]{spinde2020_integrated}
Spinde, T.; Hamborg, F.; and Gipp, B. 2020.
\newblock An Integrated Approach to Detect Media Bias in German News Articles.
\newblock In \emph{Proceedings of the ACM/IEEE Joint Conference on Digital Libraries in 2020}, JCDL '20, 505--506. Virtual Event, China: Association for Computing Machinery.
\newblock ISBN 9781450375856.

\bibitem[{Spinde et~al.(2021)Spinde, Plank, Krieger, Ruas, Gipp, and Aizawa}]{spinde-etal-2021-neural-media}
Spinde, T.; Plank, M.; Krieger, J.-D.; Ruas, T.; Gipp, B.; and Aizawa, A. 2021.
\newblock Neural Media Bias Detection Using Distant Supervision With {BABE} - Bias Annotations By Experts.
\newblock In \emph{Findings of the Association for Computational Linguistics: EMNLP 2021}, 1166--1177. Punta Cana, Dominican Republic: Association for Computational Linguistics.

\bibitem[{Thesen and Yildirim(2022)}]{electoral_thesen_2022}
Thesen, G.; and Yildirim, T.~M. 2022.
\newblock Electoral Systems and Gender Inequality in Political News: Analyzing the News Visibility of Members of Parliament in Norway and the UK.
\newblock \emph{American Political Science Review}.

\bibitem[{Wevers(2019)}]{wevers-2019-using}
Wevers, M. 2019.
\newblock Using Word Embeddings to Examine Gender Bias in {D}utch Newspapers, 1950-1990.
\newblock In \emph{Proceedings of the 1st International Workshop on Computational Approaches to Historical Language Change}, 92--97. Florence, Italy: Association for Computational Linguistics.

\bibitem[{Wu et~al.(2020)Wu, Wu, Qi, and Huang}]{wu-etal-2020-sentirec}
Wu, C.; Wu, F.; Qi, T.; and Huang, Y. 2020.
\newblock {S}enti{R}ec: Sentiment Diversity-aware Neural News Recommendation.
\newblock In \emph{Proceedings of the 1st Conference of the Asia-Pacific Chapter of the Association for Computational Linguistics and the 10th International Joint Conference on Natural Language Processing}, 44--53. Suzhou, China: Association for Computational Linguistics.

\bibitem[{Young et~al.(2021)Young, Sarnoff, Lang, and Ram\'{\i{}}rez}]{coverage_young_2021}
Young, M.; Sarnoff, H.; Lang, D.; and Ram\'{\i{}}rez, A. 2021.
\newblock Coverage and Framing of Immigration Policy in US Newspapers.
\newblock \emph{The Milbank quarterly}.

\end{thebibliography}

\appendix

\section{Paper Checklist}

\begin{enumerate}

\item For most authors...
\begin{enumerate}
    \item  Would answering this research question advance science without violating social contracts, such as violating privacy norms, perpetuating unfair profiling, exacerbating the socio-economic divide, or implying disrespect to societies or cultures?
    \answerYes{Yes, the question of improving the understanding of the media political landscape shouldn't imply any of these.}
  \item Do your main claims in the abstract and introduction accurately reflect the paper's contributions and scope?
    \answerYes{Yes}
   \item Do you clarify how the proposed methodological approach is appropriate for the claims made? 
    \answerYes{Yes}
   \item Do you clarify what are possible artifacts in the data used, given population-specific distributions?
    \answerNA{NA}
  \item Did you describe the limitations of your work?
    \answerYes{Yes, see \ref{sec:limitations}}
  \item Did you discuss any potential negative societal impacts of your work?
    \answerNo{No, because we don't think there is}
      \item Did you discuss any potential misuse of your work?
    \answerNA{NA}
    \item Did you describe steps taken to prevent or mitigate potential negative outcomes of the research, such as data and model documentation, data anonymization, responsible release, access control, and the reproducibility of findings?
    \answerNA{NA}
  \item Have you read the ethics review guidelines and ensured that your paper conforms to them?
    \answerYes{Yes}
\end{enumerate}

\item Additionally, if your study involves hypotheses testing...
\begin{enumerate}
  \item Did you clearly state the assumptions underlying all theoretical results?
    \answerNA{NA}
  \item Have you provided justifications for all theoretical results?
    \answerNA{NA}
  \item Did you discuss competing hypotheses or theories that might challenge or complement your theoretical results?
    \answerNA{NA}
  \item Have you considered alternative mechanisms or explanations that might account for the same outcomes observed in your study?
    \answerNA{NA}
  \item Did you address potential biases or limitations in your theoretical framework?
    \answerNA{NA}
  \item Have you related your theoretical results to the existing literature in social science?
    \answerNA{NA}
  \item Did you discuss the implications of your theoretical results for policy, practice, or further research in the social science domain?
    \answerNA{NA}
\end{enumerate}

\item Additionally, if you are including theoretical proofs...
\begin{enumerate}
  \item Did you state the full set of assumptions of all theoretical results?
    \answerNA{NA}
	\item Did you include complete proofs of all theoretical results?
    \answerNA{NA}
\end{enumerate}

\item Additionally, if you ran machine learning experiments...
\begin{enumerate}
  \item Did you include the code, data, and instructions needed to reproduce the main experimental results (either in the supplemental material or as a URL)?
    \answerNo{No, we will release the code upon paper acceptance. Also, we won't be able to release the dataset for copyright purposes, but we'll supply a list of urls pointing to the collected articles, as well as the preprocessing pipeline.}
  \item Did you specify all the training details (e.g., data splits, hyperparameters, how they were chosen)?
    \answerYes{Yes, in \ref{section:pipeline}}
     \item Did you report error bars (e.g., with respect to the random seed after running experiments multiple times)?
    \answerNA{We're using model training procedures that are described in other papers}
	\item Did you include the total amount of compute and the type of resources used (e.g., type of GPUs, internal cluster, or cloud provider)?
    \answerYes{Yes for the resource type and the energy consumed in \ref{tab:energy}. There's not much interest in specifying the number or the type of cluster. Each step is scalable from 1 to N gpus.}
     \item Do you justify how the proposed evaluation is sufficient and appropriate to the claims made? 
    \answerYes{Yes in \ref{sec:analysis}}
     \item Do you discuss what is ``the cost`` of misclassification and fault (in)tolerance?
    \answerNo{No, we use aggregated metrics that cancel out the effects of misclassification in NER, Entity-Linking, and Sentiment Analysis.}
  
\end{enumerate}

\item Additionally, if you are using existing assets (e.g., code, data, models) or curating/releasing new assets, \textbf{without compromising anonymity}...
\begin{enumerate}
  \item If your work uses existing assets, did you cite the creators?
    \answerYes{See \ref{section:pipeline}}
  \item Did you mention the license of the assets?
    \answerNo{No, because they are under permissive licenses}
  \item Did you include any new assets in the supplemental material or as a URL?
    \answerNA{NA}
  \item Did you discuss whether and how consent was obtained from people whose data you're using/curating?
    \answerNo{No, because we juridically don't need to}
  \item Did you discuss whether the data you are using/curating contains personally identifiable information or offensive content?
    \answerNA{NA, news data}
\item If you are curating or releasing new datasets, did you discuss how you intend to make your datasets FAIR (see \citet{fair})?
\answerNA{NA}
\item If you are curating or releasing new datasets, did you create a Datasheet for the Dataset (see \citet{gebru2021datasheets})? 
\answerNA{NA}
\end{enumerate}

\item Additionally, if you used crowdsourcing or conducted research with human subjects, \textbf{without compromising anonymity}...
\begin{enumerate}
  \item Did you include the full text of instructions given to participants and screenshots?
    \answerNA{NA}
  \item Did you describe any potential participant risks, with mentions of Institutional Review Board (IRB) approvals?
    \answerNA{NA}
  \item Did you include the estimated hourly wage paid to participants and the total amount spent on participant compensation?
    \answerNA{NA}
   \item Did you discuss how data is stored, shared, and deidentified?
   \answerNA{NA}
\end{enumerate}

\end{enumerate}

\section{Ethics Statement}
This work is part of a project which was reviewed and approved by our institution's ethical committee. 
The committee provided useful guidance concerning the selection of news sources, and the granularity of the analysis. 
The recommendations were integrated in the work done. 
 
Data were collected from publicly available news outlets, and content is not distributed directly to protect copyright.
Only the intermediate data needed for the analysis will be provided to third parties, after minimizing them in accordance to the data minimization principle of (GDPR Article 5).

News articles were collected from diversified newspapers, which lean toward different political orientations.
This and the focus on aggregated results reduce the risk of mischaracterizing any of the mentioned entities or political orientations.
Analyses such as the one proposed here contribute to making the public debate healthier and are protected by the right to freedom of expression in the European Union (EHCR Article 10). 
Moreover, all authors did their best to take an objective stance when analyzing the obtained results, and to base the claims made on reputable and objective sources.

\section{Pipeline Implementation Details}
\label{app:parameters}

\begin{table}[htbp]
    \centering
    \begin{tabularx}{\linewidth}{X|c}
        \hline
        MinHashLSH  & Parameters\\ \hline
        Word Shingle Size & 5 \\
        N\_permutations & 256 \\
        Threshold & 0.5 \\
        Library & datasketch\\
        \hline
        TSC model \\ \hline
        Base Model & \citet{martin-etal-2020-camembert} \\
        TSC-architecture & SPC~\cite{song2019attentional} \\
        Training procedure & \citet{dufraisseMADTSCMultilingualAligned2023} \\
        \hline
        BM25 \\ \hline
        Library & ElasticSearch 7.17 \\
        Index Document & title + body \\ 
        Number of Shards & 1
        \\ \hline
    \end{tabularx}
    \caption{Parameters and resources used in the news analysis framework.}
    \label{tab:my_label}
\end{table}

The amount of compute is shared as follow:

\begin{table}[h]
\centering
\begin{tabular}{|c|c|c|}
\hline
Processing Step & Hardware & Est. Power Consumption \\ \hline
NER   & Nvidia-V100   & 10 kWh   \\ \hline
Entity Linking   & Nvidia-V100   & 20 kWh  \\ \hline
TSC   & Nvidia-V100   &  2 kWh \\ \hline
\end{tabular}
\caption{Main energy consumption sources of the pipeline}
\label{tab:energy}
\end{table}

\section{Dataset Statistics}
\label{app:statistics}

\begin{figure}[htb]
	\centering
	\includegraphics[width=0.999\linewidth, trim={0cm 0cm 0cm 0cm}]{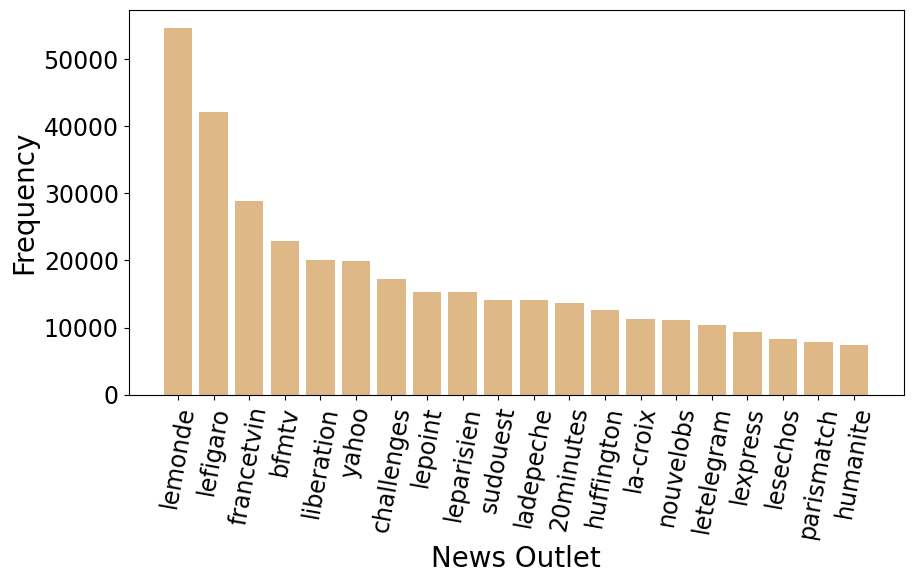} 
	\caption{Number of articles per news outlet over the 20 most frequent outlets in the corpus}
	\label{fig:dist_freq_outlet}
\end{figure}

Due to the different availability of content from news outlets, the distribution of the number of articles among them is uneven. Figure~\ref{fig:dist_freq_outlet} is the distribution of articles over the 20 most frequent news outlets.

\begin{figure}[htb]
	\centering
	\includegraphics[width=0.999\linewidth, trim={0cm 0cm 0cm 0cm}]{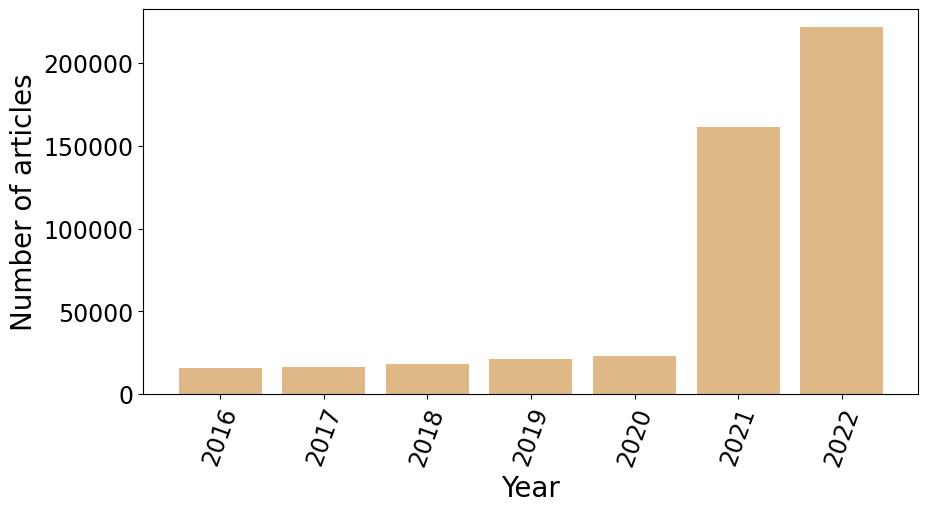} 
	\caption{Number of articles per year in the corpus}
	\label{fig:dist_year_outlet}
\end{figure}

The collection of news articles was also limited by the availability of historical articles on the media platforms. Figure~\ref{fig:dist_year_outlet} shows how uneven is the distribution. One solution in the future could be to expand the number of articles for past years using CommonCrawl snapshots. 

\begin{table}[ht]
	\begin{center}
		\resizebox{0.99\linewidth}{!}{
\begin{tabular}{|l|c|}
\hline
\multirow{1}{*}{\textbf{Split}}& Num Mentions  \\ \hline
All Persons & 2 279 463 \\ \hline
Persons Linked & 1 237 736\\ \hline
Politicians & 519 635\\ \hline
\multicolumn{2}{|c|}{Politicians Countries}\\ \hline
France & 344 903\\ \hline
Other &  174 132\\ \hline
\multicolumn{2}{|c|}{Politicians Leanings}\\ \hline
Radical-Right (RR) & 113 362 \\ \hline
Center-Right (CR) & 87 350 \\ \hline
Center (C) & 45 872 \\ \hline
Center-Left (CL) & 56 366 \\ \hline
Radical-Left (RL) & 30 484 \\ \hline
\multicolumn{2}{|c|}{Politicians per Thematic}\\ \hline
Climate & 2 114 \\ \hline
Corruption & 4 783 \\ \hline
Covid Economy & 7 381 \\ \hline
Covid Health & 8 027 \\ \hline
Yellow Vests & 2 923 \\ \hline
Immigration & 4 447 \\ \hline
Purchasing Power & 5 241 \\ \hline
Syria War & 1 511 \\ \hline
Ukraine War & 3 380 \\ \hline
Ukraine Economy & 4 286 \\ \hline
\multicolumn{2}{|c|}{Politicians Genders}\\ \hline
Man & 429 657 \\ \hline
Woman & 89 970 \\ \hline
Trans-Man & 3 380 \\ \hline
Trans-Woman & 4 286 \\ \hline

\end{tabular}			
		}
	\end{center}
	\caption{Description of dataset splits in terms of mentions}
	\label{tab:description_data}
\end{table} 

\section{Description of news sources}
\label{app:source_description}

\begin{table*}[ht]
	\begin{center}
			\begin{tabular}{|p{0.15\textwidth}|p{0.2\textwidth}|p{0.55\textwidth}|}
\hline
\textbf{Outlet} & \textbf{Website} & \textbf{Description} \\ \hline
\textit{lemonde} & \url{lemonde.fr} & Daily generalist newspaper with a centrist positioning  \\ \hline
\textit{lefigaro} & \url{lefigaro.fr} &  Daily generalist newspaper with a center-right positioning \\ \hline
\textit{francetvinfo} & \url{francetvinfo.fr} & News branch of the French national radio and TV broadcaster \\ \hline
\textit{bfmtv} & \url{bfmtv.com} & News-focused TV channel \\ \hline
\textit{liberation} & \url{liberation.fr} &  Daily generalist newspaper with a left-wing positioning \\ \hline
\textit{yahoo} & \url{yahoo.fr} & Generalist online news portal \\ \hline
\textit{challenges} & \url{challenges.fr} & Weekly business magazine with centrist views \\ \hline
\textit{lepoint} & \url{lepoint.fr} & Weekly political news magazine with centrist views \\ \hline
\textit{leparisien} & \url{leparisien.fr} & Daily generalist newspaper with a centrist positioning \\ \hline
\textit{sudouest} & \url{sudouest.fr} & Second largest regional daily newspaper in France, with a centrist positioning \\ \hline
\textit{les-crises} & \url{les-crises.fr} & Business news website considered counter-information and anti-establishment right-leaning\\ \hline
\textit{causeur} & \url{causeur.fr} & Monthly generalist newspaper with right-wing positioning\\ \hline
\textit{mediapart} & \url{mediapart.fr} & Investigative French online newspaper with left-wing positioning\\ \hline
\textit{ladepeche} & \url{ladepeche.fr} & Regional daily newspaper with a centrist positioning\\ \hline
\textit{nicematin} & \url{nicematin.com} & Regional daily newspaper with a center-right positioning\\ \hline
\textit{letelegramme} & \url{letelegramme.fr} & Regional daily newspaper with a centrist positioning\\ \hline
\textit{lesechos} & \url{lesechos.fr} & Business news daily with a centrist positioning\\ \hline
\textit{latribune} & \url{latribune.fr} & Business news weekly with a centrist positioning\\ \hline
\textit{humanite} & \url{humanite.fr} & Daily generalist newspaper with a left positioning\\ \hline
\textit{closermag} & \url{closer.fr} & Weekly press people magazine with a centrist positioning\\ \hline

\textit{valeursactuelles} & \url{valeursactuelles.com} & Weekly news magazine with a right-wing positioning\\ \hline

\end{tabular}
	\end{center}
	\caption{The table presents the outlets used in the text, their websites, and a short description.}
	\label{tab:source_description}
\end{table*} 

Table~\ref{tab:source_description} provides a brief description of top-10 outlets and the other outlets mentioned in the text. 
These descriptions are based on existing works such as~\citet{cointet2021uncovering} and \citet{lenoir2019montaigne}

\section{Representatives of the French Political Orientations}
\label{app:representatives}

\begin{table*}[ht]
	\begin{center}
			\begin{tabular}{|p{0.15\textwidth}|p{0.8\textwidth}|}
\hline
\textbf{Orientation} & \textbf{Most cited representatives (sentiment)} \\ \hline
Radical left & \ Jean-Luc Mélenchon (-0.052); Fabien Roussel (0.0920); Philippe Poutou (-0.035); Nathalie Arthaud (0.1269); François Ruffin (0.0720) \\ \hline
Center left & Anne Hidalgo (-0.09); François Hollande (-0.192); Benoît Hamon (-0.098); Ségolène Royal (-0.081); Arnaud Montebourg (-0.035); \\ \hline
Center & Emmanuel Macron (-0.074); Olivier Véran (-0.015); Jean Castex (0.0167); Gérald Darmanin (-0.091); Bruno Le Maire (0.0242) \\ \hline
Center right & Valérie Pécresse (-0.006); Nicolas Sarkozy (-0.319); François Fillon (-0.344); Éric Ciotti (0.0782); Xavier Bertrand (-0.010) \\ \hline
Radical right & Éric Zemmour (-0.176); Marine Le Pen (-0.265); Jean-Marie Le Pen (-0.201); Marion Maréchal (-0.036); Guillaume Peltier (-0.216) \\ \hline

\end{tabular}
	\end{center}
	\caption{Presentation of the five most frequently cited French representatives of each political orientation, along with associated average sentiment score.}
	\label{tab:top_per_orientation}
\end{table*} 

We complement the analysis of sentiment per political orientation from Subsection~\ref{subsec:source} in Table~\ref{tab:top_per_orientation}. 
It lists the sentiment associated with the five most cited politicians belonging to each political orientation.

\section{Representation of Mentions and Sentiment for News Outlets}
\label{app:source_ms}

\begin{figure}[htb]
	\centering
	\includegraphics[width=0.99\linewidth, trim={0cm 0cm 0cm 0cm}]{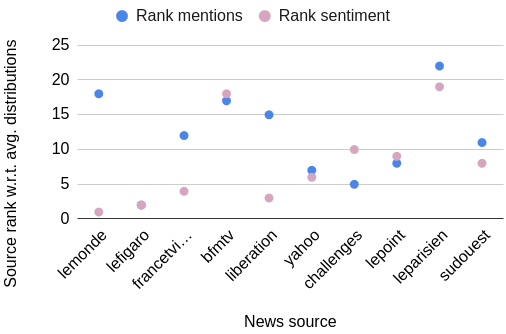} 
	\caption{Ranks of individual sources in terms of their distance to the average distributions of mentions and sentiment in the corpus. The smaller the rank is, the more similar the source is to the average. The total number of sources included in the analysis is 40. 
	}
	\label{fig:source_ms}
\end{figure}

We build vectorial representations of news outlets to capture their objective and subjective perspectives on the political spectrum.
The mentions vector stores the number of occurrences of each politician, while the sentiment vector encodes the average sentiment about each politician.
We use the 1,000  most frequently mentioned politicians as support. To obtain a reliable representation, only politicians which are mentioned at least 10 times are kept in the sentiment vectors.
Average representations of the corpus are also built and used as reference.
The individual representations are compared with average ones using cosine similarity, and ranked based on this measure. 
Figure~\ref{fig:source_ms} presents the mentions and sentiment ranks for the top 10 sources. 
The lower the rank is, the more similar the source is to the average.
For instance, \textit{lefigaro}, one of the major French daily newspapers, is ranked second for both representations,  this means that the distribution of entities and of their associated sentiment is very close to that of the entire corpus.
A large gap between the two representations is observed for \textit{lemonde}.
This source conveys a sentiment which is close to the average, but the distribution of politician mentions is rather far from the average one. 
This is also the case for \textit{liberation} and \textit{francetvinfo}, which are rather consensual in terms of sentiment conveyed, but deviate from the average for mentions of politicians.
We note that \textit{lemonde}, \textit{lefigaro} and \textit{liberation}, three of the main French newspapers are ranked first, second and third in terms of sentiment representation. 
This occurs despite their different political leanings, which are center left, center right and left, respectively~\cite{cointet2021uncovering}. 

\begin{figure}
	\centering
	\includegraphics[width=0.999\linewidth, trim={0cm 0cm 0cm 0cm}]{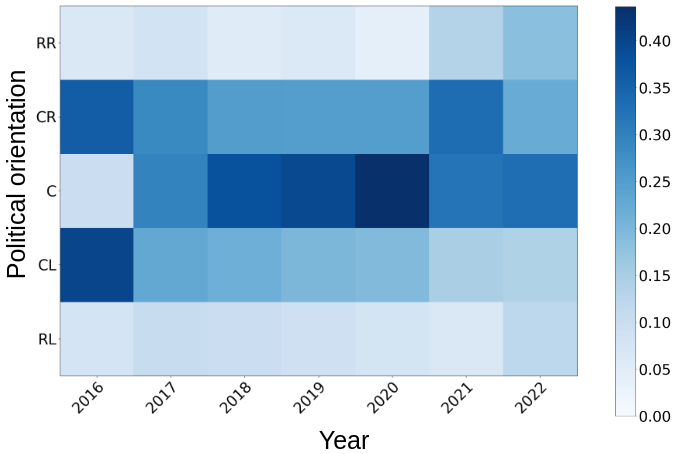} 
	\caption{Mentions of political orientations in time. }
	\label{fig:orientations_mentions_time}
\end{figure}

\begin{figure}
	\centering
	\includegraphics[width=0.999\linewidth, trim={0cm 0cm 0cm 0cm}]{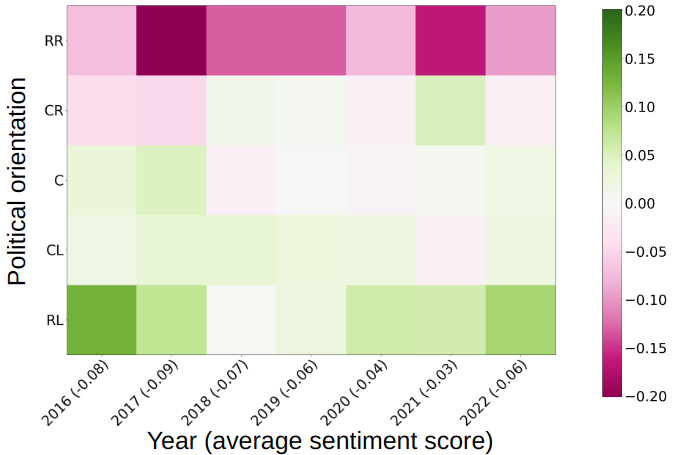} 
	\caption{Sentiment associated with political orientations in time. }
	\label{fig:orientations_sentiment_time}
\end{figure}

\section{Temporal Dynamics of Politicians}
\label{app:temporal_politicians}

\begin{figure}
	\centering
	\includegraphics[width=0.999\linewidth, trim={0cm 0cm 0cm 0cm}]{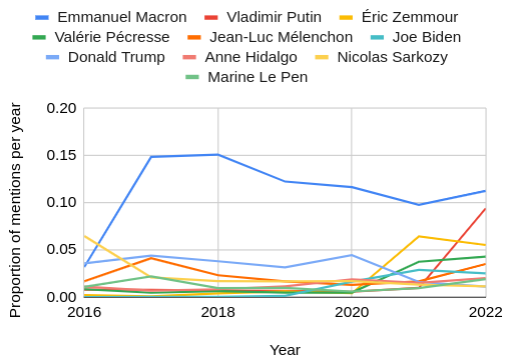} 
	\caption{Mentions of most frequently mentioned politicians in time. }
	\label{fig:top_mentions_time}
\end{figure}

\begin{figure}
	\centering
	\includegraphics[width=0.999\linewidth, trim={0cm 0cm 0cm 0cm}]{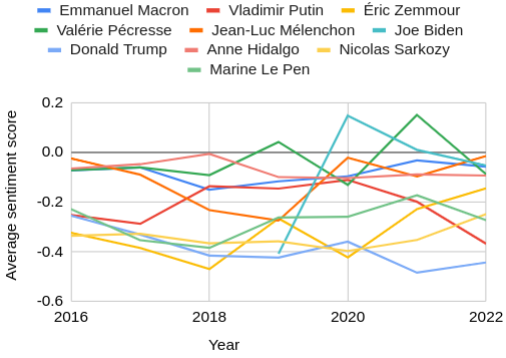} 
	\caption{Sentiment associated with the most frequently mentioned politicians in time. }
	\label{fig:top_sentiment_time}
\end{figure}
We present the temporal evolution of mentions of the ten most frequently mentioned politicians in Figures~\ref{fig:top_mentions_time} and~\ref{fig:top_sentiment_time}.
The detailed view of mentions from Figure~\ref{fig:top_mentions_time} illustrates the strong representation of \textit{Emmanuel Macron} in the corpus. 
We also note the surge of mentions of \textit{Vladimir Putin} in 2022, and that for \textit{Eric Zemmour} in 2021.
Sentiment changes in variable proportions in Figure~\ref{fig:top_sentiment_time}. 
Stronger variations are explained by different factors. 
\textit{Joe Biden} was initially absent from the corpus, then was presented negatively, but sentiment became more positive when his 2020 candidacy gained traction and he won the US presidential elections;
\textit{Eric Zemmour} was portrayed very negatively prior to 2021, but his image improved for some news outlets when his 2022 presidential run gained momentum; 
\textit{Vladimir Putin} has an overall negative representation that has worsened since the invasion of Ukraine. 
Weaker variations are observed for \textit{Emmanuel Macron} whose sentiment score is close to the average from Figure~\ref{fig:source_sentiment}, except 2018, the year of the \textit{Yellow Vests} movement, when his image was degraded.
\textit{Nicolas Sarkozy} has a constantly negative score which is probably explained by the array of legal actions opened against him.

\section{Temporal Dynamics of Political Orientations}
\label{app:temporal_orientations}

Figures~\ref{fig:orientations_mentions_time} and~\ref{fig:orientations_sentiment_time} illustrate the temporal evolution of mentions and average sentiment associated with political orientations. 
This analysis complements the outlet-oriented one from Subsection~\ref{subsec:source}. 
The distribution of mentions changes strongly between 2016 and 2017.
2016 is dominated by the center-left, which included the governing party at the time, and the center right.
The central orientation gains momentum in 2017, when  its party became majoritary, and is dominant between 2018 and 2020.
The center right is well represented in 2021, when its primary elections for the  2022 presidential election took place and were frequently discussed in the media, with the center also being frequently mentioned.
Then, the center again becomes dominant in 2022. 
The representation of the radical right  became stronger in 2021 and 2022, when \textit{Marine Le Pen} and \textit{Eric Zemmour} played important roles in the campaign for the presidentioal election.
A similar trend is observed for the radical left, whose main representative, \textit{Jean-Luc Mélenchon} was also well placed in the presidential race.

\begin{figure}[htb]
	\centering
	\includegraphics[width=0.99\linewidth, trim={0cm 0cm 0cm 0cm}]{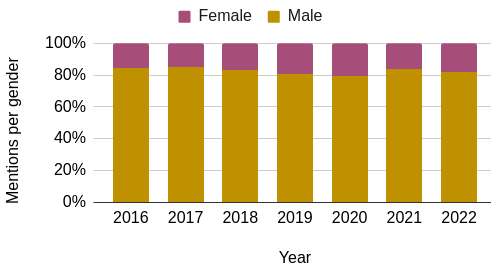} 
	\caption{Percentage of gender mentions per year.
	}
	\label{fig:gender_years}
\end{figure}

The evolution of sentiment is presented in Figure~\ref{fig:orientations_sentiment_time}.
It confirms that left-wing orientations have a media coverage which is overall more favorable compared to that of their right counterparts. 
The contrast is most marked for the radical orientations, with RL sentiment being positive during most years, and particularly in 2016 and 2022.
Inversely, the sentiment toward the radical right is more negative in 2017, when \textit{Marine Le Pen} lost the presidential election, and in 2021. 
The sentiment toward the center was most positive in 2017, the year when \textit{Emmanuel Macron} won his first presidential run. 
However, this sentiment became negative in 2018, the year when the \textit{Yellow Vests} movement errupted. 

\section{Temporal Dynamics of Gender Representation}
\label{app:temporal_gender}

The gender-oriented analysis presented in Subsection~\ref{subsec:gender} is enriched with a presentation in Figure~\ref{fig:gender_years} of the evolution of gender mentions in time.
There is overall little variation of the gender imbalance over the studied period.
A small reduction of this bias is noted between 2017 and 2020, but the proportion of female mentions decreases in 2020. 

\end{document}